\documentclass[letterpaper, 10 pt, conference]{ieeeconf}  % Comment this line out if you need a4paper

\IEEEoverridecommandlockouts                              % This command is only needed if 
\usepackage{amsmath, mleftright}    % need for subequations
\usepackage{graphicx}   % need for 
\usepackage[lofdepth,lotdepth]{subfig}
\usepackage{tabularx}
\usepackage[font=footnotesize,labelfont=footnotesize]{caption}
\usepackage{verbatim}   % useful for program listings
\usepackage{color}      % use if color is used in text
\usepackage{hyperref} % use for hypertext links, including those to external documents and URLs
\usepackage{multirow}
\usepackage{multicol}
\usepackage{graphicx}
\usepackage{rotating}
\usepackage{array}
\usepackage{xspace}
\usepackage{indentfirst}
\usepackage{amssymb}
\usepackage{float}
\usepackage{algorithm}
\usepackage[algo2e]{algorithm2e}
\usepackage{algorithmic}
\usepackage{comment}
\usepackage{sidecap}
\usepackage{booktabs}
\usepackage{breqn}
\usepackage{cite}
\usepackage{mathtools}
\usepackage{graphicx}
\graphicspath{{./figures/}}
\usepackage{caption}
\usepackage{float}
\usepackage{lipsum}
\usepackage{xcolor}
\usepackage{siunitx}
\usepackage{xspace}
\usepackage{bm}
\usepackage{gensymb}
\let\labelindent\relax
\usepackage[inline]{enumitem}
\usepackage{paralist}
\usepackage{graphicx}

\newcommand\norm[1]{\left\lVert#1\right\rVert}
\DeclareMathOperator*{\argmin}{argmin} % thin space, limits underneath in displays

\newcommand{\gobble}[1]{}

\newcounter{tecounter}
\title{\LARGE \bf
Target-free Extrinsic Calibration of a 3D-Lidar and an IMU
}

\author{Subodh Mishra$^{1}$, Gaurav Pandey$^{2}$ and Srikanth Saripalli$^{1}$% <-this % stops a space
\thanks{$^{1}$ with the Department of Mechanical Engineering, Texas A\&M University
        {\tt\small subodh514@tamu.edu}}%
\thanks{$^{2}$ with the Ford Motor Company, USA}
\thanks{The authors would like to thank the Ford Motor Company for financially supporting this work.}
}

\begin{document}
\maketitle
\thispagestyle{empty}
\pagestyle{empty}

%%%%%%%%%%%%%%%%%%%%%%%%%%%%%%%%%%%%%%%%%%%%%%%%%%%%%%%%%%%%%%%%%%%%%%%%%%%%%%%%
\begin{abstract}
This work presents a novel target-free extrinsic calibration algorithm for a 3D Lidar and an IMU pair using an Extended Kalman Filter (EKF) which exploits the \textit{motion based calibration constraint} for state update. The steps include, data collection by motion excitation of the Lidar Inertial Sensor suite along all degrees of freedom, determination of the inter sensor rotation by using rotational component of the aforementioned \textit{motion based calibration constraint} in a least squares optimization framework, and finally, the determination of inter sensor translation using the \textit{motion based calibration constraint} for state update in an Extended Kalman Filter (EKF) framework. We experimentally validate our method using data collected in our lab and open-source (\url{https://github.com/unmannedlab/imu_lidar_calibration}) our contribution for the robotics research community.
\end{abstract}

\begin{keywords}
Extrinsic Calibration, Lidar, IMU, Optimization, Extended Kalman Filter
\end{keywords}
%%%%%%%%%%%%%%%%%%%%%%%%%%%%%%%%%%%%%%%%%
\section{Introduction} 
\label{sec: introduction}
3D-Lidars and IMUs are ubiquitous to autonomous robots. 3D-Lidars provide a 3D point cloud of the area that the robot operates in and are not affected by illumination. This has proved the usage of Lidars beneficial in several robotic applications. However, owing to the spinning nature of the sensor and the sequential manner in which they produce measurements, Lidars suffer from significant motion distortion when a robot exhibits dynamic maneuvers. The motion distortion can be seen in highway operating speeds for self driving cars and also in other robotic applications like autonomous flight and off-road robotics. Motion distorted scans deteriorate the result of Lidar Odometry/Simultaneous Localizaton $\&$ Mapping (SLAM) algorithms.

Inertial Measurement Units (IMUs) can be used to mitigate the effect of motion on spinning Lidars. IMUs measure linear acceleration and angular velocity at frequencies higher than the spinning rate of a Lidar. State of the art Lidar Odometry/SLAM algorithms use IMUs to correct motion distortion (also called deskewing) in Lidar scans and produce better estimates of the robot's position and the surrounding map. In order to use the IMU's measurements, these algorthms require that the spatial separation or the extrinsic calibration between the IMU and Lidar be known \textit{a-priori}, such that data from both these sensors can be expressed in a common frame of reference. In robotics labs where researchers generally assemble sensor suites using products procured from different sources, the extrinsic calibration between a Lidar and an IMU is usually unkown. Therefore, it is important to estimate the extrinsic calibration in order to use any Lidar Inertial Odometry or SLAM algorithm. Although IMUs operate at frequencies much higher than the spinning rate of Lidars, the number of times a Lidar fires a beam to acquire measurements during a 360 $^{\circ}$ scan (\emph{viz,} the firing rate) is significantly higher than the IMU frequency. This requires the use of interpolation/extrapolation techniques to match IMU rates to Lidar firing rates so that motion compensation can be done using IMU measurements. \cite{lincalib1} uses Gaussian process regression, \cite{lincalib2} uses continuous time splines, while we use a discrete time IMU state propagation model under an EKF framework to compensate for the effect of motion during the calibration process.
\begin{figure}[]
    \centering
    \includegraphics[width=0.45\textwidth]{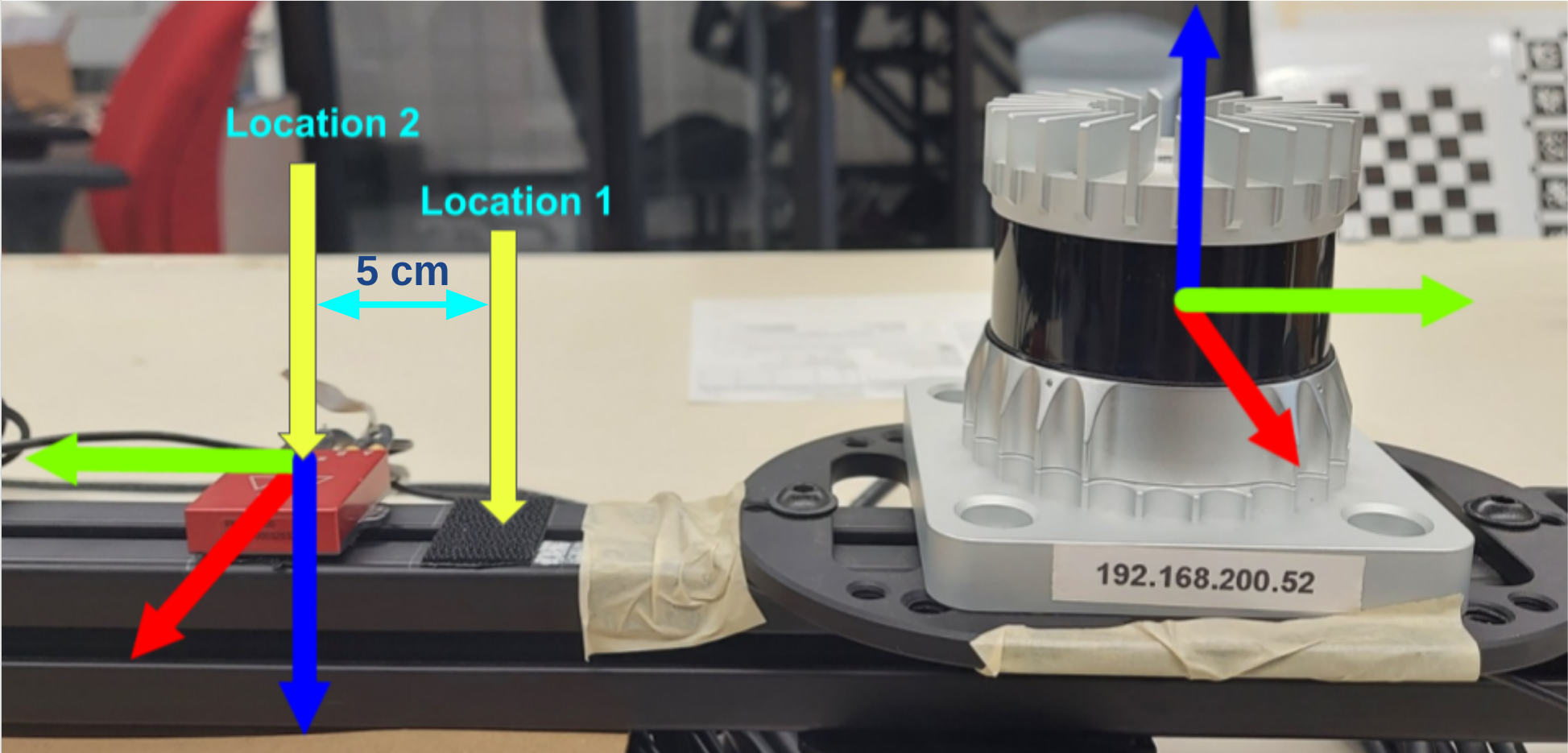}
    \caption{\textbf{Experimental Platform: } Lidar Inertial Sensor Suite with an Ouster 128 Channel Lidar and a Vectornav-VN 300 IMU. Red: x axis, Green: y-axis, Blue: z-axis. We place the IMU at two different locations (highlighted by yellow arrows), which, according to ruler measurements, are apart from each other by 5 cm along the y axis. In the absence of ground truth, we will use this information as a reference to validate our calibration algorithm.}
    \label{fig:LidarIMUSystem}
    \vspace{-20pt}
\end{figure}
\section{Related work}
\label{sec: relatedwork}
There are several published works on extrinsic calibration of 3D-Lidar camera systems (e.g. \cite{ppccal}, \cite{pbpccal}, \cite{msgcal}, \cite{jiang2021calibrating}) and camera IMU systems (e.g. \cite{KalmanFilterBasedCamCalib}, \cite{kalibr}, \cite{kalibr2}) which do not depend on a third auxiliary sensor. There are numerous published works on 3D-Lidar IMU calibration as well \emph{viz.} \cite{lincalib3}, \cite{lincalib4}, \cite{lincalib5}, \cite{lincalib6}, \cite{lincalib1}, \cite{lincalib2}. However, among these, \cite{lincalib3}, \cite{lincalib4}, \cite{lincalib5} are methods which use GPS/GNSS information for accurate pose estimation of the inertial sensor. Therefore, these approaches cannot be used when GPS is unavailable. \cite{lincalib6} does camera IMU calibration first and then cross calibrates the visual inertial system with the Lidar. \cite{lincalib1}, \cite{lincalib2} do not depend on auxilary sensors for 3D-Lidar IMU extrinsic calibration, thus making them easier to use in cases when an auxiliary sensor is unavailable.

\cite{lincalib1} uses Gaussian Process regression  \cite{GPR} to up-sample IMU measurements so that a corresponding IMU reading can be inferred for every Lidar firing time. Upsampling of IMU measurements helps remove motion distortion from Lidar scans which occurs during the data collection phase of Lidar imu calibration. They use 3 orthogonal planes as a calibration reference map/target and utilize the projection of each measured Lidar point on the corresponding plane as a geometrical constraint which requires the knowledge of the unkown extrinsic calibration parameters. They perform IMU-pre-integration \cite{DBLP:journals/corr/ForsterCDS15} on the up-sampled IMU measurements and solve a non linear batch estimation problem to determine the unknown extrinsic calibration parameters. 

\cite{lincalib2} on the other hand models the IMU state (pose, velocity, biases), instead of IMU measurements (like \cite{lincalib1}), as a continuous time spline which can be differentiated and equated to IMU measurements. Modelling the IMU trajectory as a continous time spline helps infer IMU pose at Lidar firing times. Unlike the previous method (i.e. \cite{lincalib1}), this method exploits all the planes in the calibration environment and similar to the previous method this method also uses the projection of each measured Lidar point on the corresponding plane as a geometrical constraint which requires the knowledge of the unknown extrinsic calibration parameters. The geometrical constraint is squared and added to form a cost function which is minimized using the method of non linear least squares. Although \cite{lincalib2}  does not use any calibration target, it nonetheless requires an environment which needs to have several well defined planar structures.

Our approach, which utilizes an Extended Kalman Filter (EKF) for 3D-Lidar IMU extrinsic calibration, draws inspiration from the camera IMU extrinsic calibration algorithm presented in \cite{KalmanFilterBasedCamCalib}. In contrast to \cite{KalmanFilterBasedCamCalib}, we use a measurement model suitable for Lidar. In contrast to \cite{lincalib1} and \cite{lincalib2}, which use Gaussian process regression and splines respectively, we use a discrete time IMU state propagation model (Equations \ref{eqn: propeqn1} - \ref{eqn: propeqn7}) under an EKF framework to remove motion distortion in Lidar scans.

Extrinsic calibration is required to remove motion distortion and motion distortion occurs while collecting data to do extrinsic calibration. The EKF framework allows us to use the best available estimate of extrinsic calibration parameters to motion compensate the Lidar scans during the extrinsic calibration process. Our approach follows a predict (Section \ref{sec: stateprop}) $\longrightarrow$ deskew (Section \ref{sec: deskewing}) $\longrightarrow$ update (Section \ref{sec: stateupdate}) cycle, with a scan matching (Section \ref{sec: ndtscanmatching}) technique which uses the deskewed scans to produce measurements for state/covariance update (Figure \ref{fig:KFBlockDiagran}) in order to do extrinsic calibration of a 3D-Lidar IMU system. We use the OpenVINS\cite{OpenVins} framework to develop our calibration code. 

\section{Contributions}
The proposed approach does not require any calibration target (unlike \cite{lincalib1}) or specific environmental features like planes (unlike both \cite{lincalib1} and \cite{lincalib2}) for doing extrinsic calibration of a 3D-Lidar and an IMU. Our 3D-Lidar IMU extrinsic calibration algorithm does not depend on usage of auxiliary pose sensors like GPS/GNSS or camera-inertial sensor suites, which \cite{lincalib3}, \cite{lincalib4}, \cite{lincalib5}, \cite{lincalib6} do. Our method does not require any tape measured initialization of the calibration parameters (unlike \cite{KalmanFilterBasedCamCalib}). We utilize a generic EKF for Visual Inertial Navigation presented in \cite{OpenVins} and re-purpose it to incorporate the motion based calibration constraint utilized in \cite{mocalZachTaylor} to formulate our own 3D-Lidar IMU extrinsic calibration algorithm. Finally, to the best of our knowledge, this contribution is the second open-sourced 3D-Lidar IMU calibration algorithm which does not depend on any auxiliary sensor, with \cite{lincalib2} being the first.
\section{Motion based extrinsic calibration}
\label{sec: mocal}

\begin{figure}[!ht]
    \centering
    \includegraphics[scale=0.20]{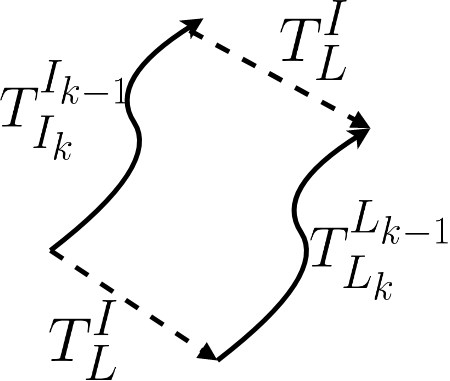}
    \caption{Motion based calibration constraint}
    \label{fig:motioncalibration}
\end{figure}

The motion based constraint for corresponding sensor motion is given in Equation \ref{eqn: fullmotionconstraint} and a schematic is shown in Figure \ref{fig:motioncalibration}. This constraint is similar to the standard Hand Eye Calibration \cite{hec} constraint commonly used in calibration of robotic manipulators and extended to calibration of multimodal mobile robotic sensors in \cite{mocalZachTaylor}. In our scenario, $T^{L_{k-1}}_{L_{k}}$ is the motion between two Lidar scans which can be obtained by using any scan matching technique \cite{NDT} and $T^{I_{k-1}}_{I_{k}}$ is the motion experienced by the IMU between the two scan instants. As the sensors are spatially separated from each other by a fixed rigid pose $T^{I}_{L}$ (the extrinsic calibration parameter), we obtain the constraint given in Equation \ref{eqn: fullmotionconstraint}. 
\begin{align}
    T^{I_{k-1}}_{I_{k}} T^{I}_{L} &= T^{I}_{L} T^{L_{k-1}}_{L_{k}} \label{eqn: fullmotionconstraint} 
\end{align}
Here, 
$T^{I_{k-1}}_{I_{k}} = 
\begin{bmatrix} R^{I_{k-1}}_{I_{k}} & ^{I_{k-1}}p_{I_{k}} \\
                              0  & 1
\end{bmatrix}$ \\ 
$T^{L_{k-1}}_{L_{k}} = 
\begin{bmatrix} R^{L_{k-1}}_{L_{k}} & ^{L_{k-1}}p_{L_{k}} \\
                              0  & 1
\end{bmatrix}$ and 
$T^{I}_{L} = 
\begin{bmatrix} R^{I}_{L} & ^{I}p_{L} \\
                              0  & 1
\end{bmatrix}$. \\
$R^{a}_{b} \in SO(3)$ is a rotation matrix and $^{a}p_{b} \in R^{3 \times 1}$ is a translation vector. We use the motion based constraint given by Equation \ref{eqn: fullmotionconstraint} for estimating the inter sensor rotation $R^{I}_{L} \in SO(3)$ and translation $^{I}p_{L} \in R^{3}$. Specifically, we utilize the rotation component of Equation \ref{eqn: fullmotionconstraint} to initialize the inter sensor rotation and use this initialization to estimate the inter sensor translation using an Extended Kalman Filter based algorithm\cite{KalmanFilterBasedCamCalib}, \cite{OpenVins}.

\section{Problem Formulation}
\label{sec: problemformulation} 

Our goal is to determine the spatial 6 DoF separation, \emph{viz.} the extrinsic calibration $T^{I}_{L} \in SE(3)$ between a 3D-Lidar and an IMU (Figure \ref{fig:LidarIMUSystem}). We divide the calibration process into three steps \emph{viz} data collection (Section \ref{sec: datacollection}), inter sensor rotation initialization (Section \ref{sec: rotationestimation}, Figure \ref{fig:rotationHEC}) and full extrinsic calibration (Section \ref{sec: fullstateestimationusingEKF}, Figure \ref{fig:KFBlockDiagran}). 

\subsection{Data Collection} 
\label{sec: datacollection} 
\begin{figure}[!ht]
    \centering
    \includegraphics[width=0.35\textwidth]{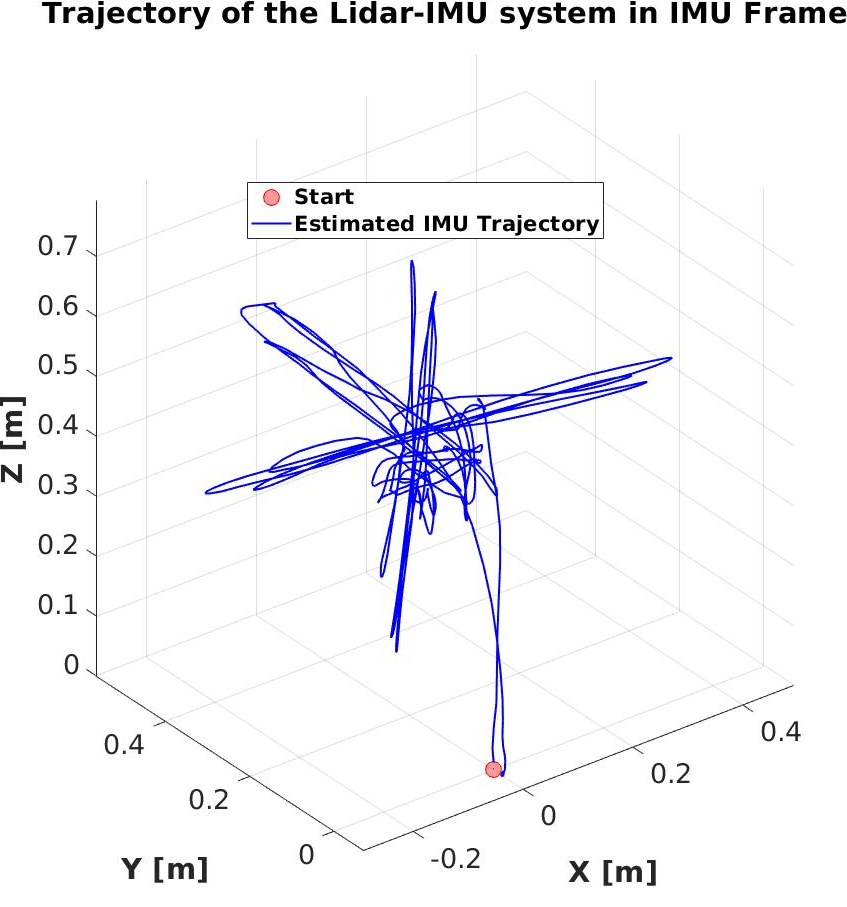}
    \caption{Trajectory of the Lidar-IMU system}
    \label{fig: Lidarimutraj}
\end{figure}

Data collection is an important step in any extrinsic calibration algorithm. Since our sensor suite involves an IMU, a proprioceptive sensor which can sense only motion, we sufficiently excite \footnote{\url{https://youtu.be/2IX5LVTDkLc}} all degrees of rotation and translation so that all the components of the extrinsic calibration parameter are completely observable. However, as remarked in \cite{KalmanFilterBasedCamCalib} motion excitation along at least two degrees of rotational freedom is essential for observability. This is demonstrated in Section \ref{sec: motionandconvergence}. Although \cite{KalmanFilterBasedCamCalib} deals in Camera IMU calibration, the remark about observability will also extend to the 3D-Lidar IMU calibration problem because in both the cases the exteroceptive sensor (camera in \cite{KalmanFilterBasedCamCalib} and Lidar in our case) is used as a pose sensor.  The data collection XYZ trajectory is shown in Figure \ref{fig: Lidarimutraj}. For the purpose of notation, let us denote that we collect $M$ Lidar scans in the process of collecting data.

\vspace{-15pt}
\subsection{Inter-sensor Rotation Initialization}
\label{sec: rotationestimation}
We estimate the rotation between the IMU and 3D-Lidar by using the rotation component of the motion based calibration constraint (Equation \ref{eqn: fullmotionconstraint}). The rotation component is given in Equation \ref{eqn: rotationmotionconstraint}.
\begin{align}
    R^{I_{k-1}}_{I_{k}} R^{I}_{L} &= R^{I}_{L} R^{L_{k-1}}_{L_{k}} \label{eqn: rotationmotionconstraint} 
\end{align}

\begin{figure*}[!ht]
  \centering
  \subfloat[Initialization of Rotation $R^{I}_{L}$]{\includegraphics[width=0.3\textwidth]{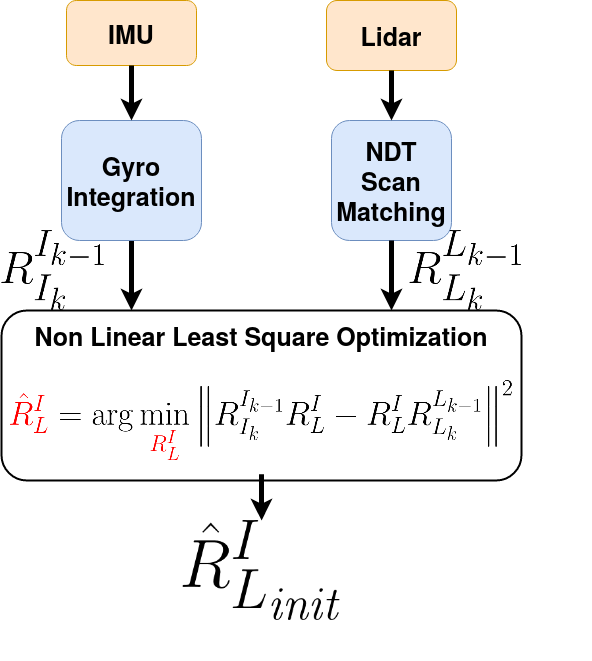}\label{fig:rotationHEC}}
  \quad
  \subfloat[Extended Kalman Filter for estimating both $R^{I}_{L}$ and $^{I}p_{L}$]{\includegraphics[width=0.65\textwidth]{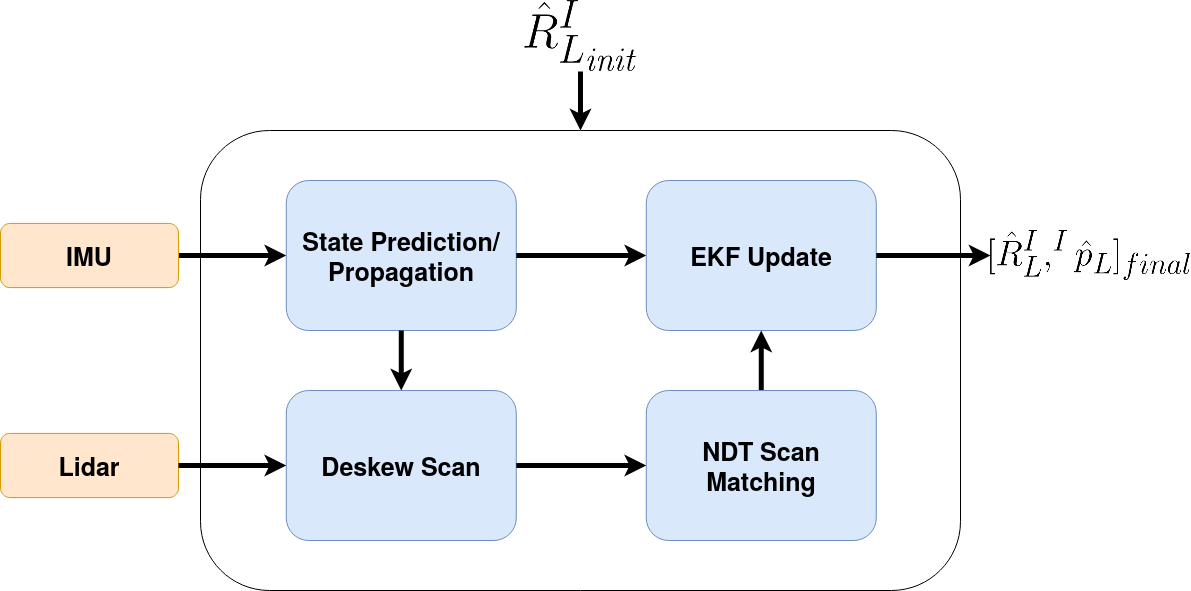}\label{fig:KFBlockDiagran}}
  \caption{Figure \ref{fig:rotationHEC} presents the process of determination of an initial estimate of $R^{I}_{L}$, Figure \ref{fig:KFBlockDiagran} presents the Extended Kalman Filter which utilizes the initial estimate of $R^{I}_{L}$ obtained in Figure \ref{fig:rotationHEC} to generate both $R^{I}_{L}$ and $^{I}p_{L}$ together.} 
  \label{fig:InitBlockKFBlock}
  \vspace{-15pt}
\end{figure*}

In our implementation we use an axis angle representation of the sensor rotations. Using the axis angle representation, Equation \ref{eqn: rotationmotionconstraint} can be reformulated as
\begin{align}
     ^{I_{k-1}}r_{I_{k}}  &= R^{I}_{L} {}^{L_{k-1}}r_{L_{k}} \label{eqn: rotationmotionconstraintAxisAngle} 
\end{align}

Here $^{I_{k-1}}r_{I_{k}}$ $\&$ $^{L_{k-1}}r_{L_{k}} \in R^{3}$ are axis angle representations for $R^{I_{k-1}}_{I_{k}}$ $\&$ $R^{L_{k-1}}_{L_{k}}$ respectively. As shown in Figure \ref{fig:rotationHEC}, we use NDT scan matching \cite{NDT} to estimate the Lidar rotation $R^{L_{k-1}}_{L_{k}}$ between consecutive Lidar scans. We integrate gyroscope measurements between two scan instants to estimate IMU rotation $R^{I_{k-1}}_{I_{k}}$. An objective function (Equation \ref{eqn: objectivefunction}), unknown in $R^{I}_{L}$, is formed by squaring and summing the constraint given in Equation \ref{eqn: rotationmotionconstraintAxisAngle} for each ${}^{L_{k-1}}r_{L_{k}}$ and the corresponding ${}^{I_{k-1}}r_{I_{k}}$. 
\begin{align}
\label{eqn: objectivefunction}
    P = \sum_{i=1}^{M-1} \norm{{}^{I_{k-1}}r_{I_{k}}  - R^{I}_{L} {}^{L_{k-1}}r_{L_{k}}}^{2}
\end{align}
In order to obtain an estimate $\hat{R}^{I}_L$, Equation \ref{eqn: objectivefunction} needs to be minimized with respect to $R^{I}_L$ as shown in Equation \ref{eqn: CostFnMinimization}. We use the Ceres \cite{ceres-solver} non linear least square solver to solve the optimization problem in Equation \ref{eqn: CostFnMinimization}.
\begin{equation}
\label{eqn: CostFnMinimization}
    \hat{R}^{I}_L = \argmin_{R^{I}_L} P
\end{equation}
In this step, the sensor rotations $R^{I_{k-1}}_{I_{k}}$ and $R^{L_{k-1}}_{L_{k}}$ used to estimate $R^{I}_{L}$ have certain shortcomings. First, $R^{I_{k-1}}_{I_{k}}$ is calculated by integrating the gyro measurements without taking the gyroscope bias into account and second, $R^{L_{k-1}}_{L_{k}}$ is obtained from NDT scan matching of Lidar scans which may have significant motion distortion due to the motion undertaken by the sensor suite during the data collection step (Section \ref{sec: datacollection}). This step only provides an initial estimate of $R^{I}_{L}$ which is used for initializing the EKF based algorithm described in Section \ref{sec: fullstateestimationusingEKF}.
\vspace{-15pt}
\subsection{Inter-sensor Translation Estimation: Full State Estimation using an Extended Kalman Filter}
\label{sec: fullstateestimationusingEKF}
The estimation of inter sensor translation ${}^{I}p_{L}$ depends on IMU translation ${}^{I_{k-1}}p_{I_{k}}$ (Equation \ref{eqn: fullmotionconstraint}) which involves double integration of IMU accelerometer measurements, but performing double integration without the knowledge of biases will introduce significant errors. We use an Extended Kalman Filter (EKF) which, in addition to estimating $R^{I}_{L}$ $\&$ $^{I}p_{L}$, also estimates the accelerometer $\&$ gyroscope biases, the pose $\&$ velocity of the IMU at the Lidar scan instants. The block diagram for the EKF approach is shown in Figure \ref{fig:KFBlockDiagran}.
% The EKF based extrinsic calibration procedure involves a predict (Section \ref{sec: stateprop}) $\longrightarrow$ deskew (Section \ref{sec: deskewing}) $\longrightarrow$ update (Section \ref{sec: stateupdate}) cycle, with a scan matching (Section \ref{sec: ndtscanmatching}) technique utilizing the deskewed scans to produce measurements for state/covariance update. 
The states we will be estimating are: 
\begin{equation}
    \mathcal{X} = \{ {X^{G}_{I_{k=0:M-1}}}, T^{I}_{L} \}
\end{equation}
Here $M$ is the number of scans. $X^{G}_{I_k}$ is the IMU state at scan timestamp $k$. Our EKF state vector has an evolving component and a static component. The evolving component is the IMU state at scan timestamp $k$:
\begin{equation}
    \hat{X}^{G}_{I_k} = \{^{I_{k}} _{G}\hat{\Bar{q}}, ^{G}\hat{\textbf{v}}_{I_{k}}, ^{G}\hat{\textbf{p}}_{I_{k}}, \hat{\textbf{b}}_{g, k}, \hat{\textbf{b}}_{a, k}\}
    \label{eqn: evolvingstatevector}
\end{equation}
$^{I_{k}} _{G}\hat{\Bar{q}}$ is the unit quaternion which encodes the IMU orientation such that the rotation matrix $R^{\top}(^{I_{k}} _{G}\hat{\Bar{q}})$ is the IMU orientation with respect to the global frame $G$. $^{G}\hat{\textbf{p}}_{I_{k}}$ $\&$ $^{G}\hat{\textbf{v}}_{I_{k}}$ are IMU position and velocity vectors ($\in R^{3\times1}$) respectively in frame $G$. $\hat{\textbf{b}}_{a,k}$ $\&$ $\hat{\textbf{b}}_{g,k}$ are the accelerometer and gyro bias vectors ($\in R^{3 \times 1}$) respectively. The static component of the EKF state vector is the extrinsic calibration, parameterized as $T^{I}_{L}$. $T^{I}_{L}$ is formed by rotation matrix $R^{I}_{L}$ and translation vector $^{I}\textbf{p}_{L}$, however in the EKF formulation we parameterize the rotation $R^{I}_{L}$ as a unit quaternion $^{I}_{L}\bar{q}$.  

\subsubsection{State Propagation}
\label{sec: stateprop} 
We use the discrete time implementation given in \cite{KalmanFilterBasedCamCalib}, \cite{OpenVins} to propagate the EKF state (Equations \ref{eqn: propeqn1} - \ref{eqn: propeqn7}) from imu timestamp $i$ to $i+1$. The gyroscope and accelerometer measurements $\bm{\omega}_{m,i} \in R^{3\times1}$ $\&$ $\textbf{a}_{m,i} \in R^{3\times1}$ respectively, are assumed to be constant during the IMU sampling period $\Delta t$. In the following equations ${}^{G}\textbf{g}$ is the acceleration due to gravity in the global frame G.
\begin{align}
    ^{I_{i+1}} _{G}\hat{\Bar{q}} &= \exp \bigg(\frac{1}{2}\Omega(\bm{\omega}_{m,i} - \hat{\textbf{b}}_{g,i})\Delta t\bigg) {}^{I_{i}} _{G}\hat{\Bar{q}} \label{eqn: propeqn1}\\
    ^{G}\hat{\textbf{v}}_{I_{i+1}} &= ^{G}\hat{\textbf{v}}_{I_{i}} - {}^{G} \textbf{g} \Delta t + \hat{\textbf{R}}^{G} _{I_{i}} (\textbf{a}_{m, i} - \hat{\textbf{b}}_{a, i}) \Delta t \label{eqn: propeqn2}\\
    ^{G}\hat{\textbf{p}}_{I_{i+1}} &= ^{G}\hat{\textbf{p}}_{I_{i}} + ^{G}\hat{\textbf{v}}_{I_{i}} \Delta t - \frac{1}{2} {}^{G}\textbf{g} \Delta t ^{2} \nonumber \\&+ \frac{1}{2} \hat{\textbf{R}}^{G} _{I_{i}} (\textbf{a}_{m, i} - \hat{\textbf{b}}_{a, i}) \Delta t^{2} \label{eqn: propeqn3} \\
    \hat{\textbf{b}}_{g, i+1} &= \hat{\textbf{b}}_{g, i} \label{eqn: propeqn4} \\
    \hat{\textbf{b}}_{a, i+1} &= \hat{\textbf{b}}_{a, i} \label{eqn: propeqn5} \\
    ^{I}_{L}\hat{\Bar{q}}_{i+1} &= ^{I}_{L}\hat{\Bar{q}}_{i} \label{eqn: propeqn6} \\
    ^{I}\hat{\textbf{p}}_{L, i+1} &= ^{I}\hat{\textbf{p}}_{L, i} \label{eqn: propeqn7}
\end{align}
In Equation \ref{eqn: propeqn1}, $\exp()$ is matrix exponential (Equation 96 in \cite{Trawny2005IndirectKF}), $\Omega(\bm{\omega}) = \begin{bmatrix}
-[\bm{\omega}]_{\times} & \bm{\omega} \\
-\bm{\omega}^{\top} & 0
\end{bmatrix}$ $\&$ $[\bm{\omega}]_{\times} = \begin{bmatrix}
0 & -\omega_{z} & \omega_{y} \\
\omega_{z} & 0 & -\omega_{x} \\
-\omega_{y} & \omega_{x} & 0 \\
\end{bmatrix}$.
The gyroscope and accelerometer measurements $\bm{\omega}_{m,i}$ and $\textbf{a}_{m,i}$ respectively, used to propagate the evolving state (Equation \ref{eqn: propeqn1}-\ref{eqn: propeqn7}) state are modelled as:
\begin{align}
    \bm{\omega}_{m} &= \bm{\omega} + \textbf{b}_{g} + \textbf{n}_{g} \nonumber\\
    \textbf{a}_{m} &= \textbf{a} + \textbf{R}^{I}_{G} {}^{G}\textbf{g} + \textbf{b}_{a} + \textbf{n}_{a}
    \label{eqn: imumodelcontinous}
\end{align}
Here $\textbf{n}_{g}$ $\&$ $\textbf{n}_{a}$ are white Gaussian noise. Discretizing and taking expected value, Equation \ref{eqn: imumodelcontinous} can be written as:
\begin{align}
    \bm{\omega}_{m, i} &= \hat{\bm{\omega}}_{i} + \hat{\textbf{b}}_{g, i} \nonumber \\
    \textbf{a}_{m, i} &= \hat{\textbf{a}}_{i} + \hat{\textbf{R}}^{I_{i}}_{G} {}^{G}\textbf{g} + \hat{\textbf{b}}_{a,i}
    \label{eqn: imumodeldiscrete}
\end{align}

In addition to propagation of state variables, we also need to propagate the EKF state covariance $\textbf{P}$ from imu timestamp $i$ to $i+1$ using Equation \ref{eqn: propcov}.
\begin{align}
    \textbf{P}_{i+1} = \Phi(t_{i+1}, t_i)\textbf{P}_{i}\Phi(t_{i+1}, t_i)^{T} + \textbf{G}_i \textbf{Q}_d \textbf{G}^{T}_i \label{eqn: propcov}
\end{align}
Here,
\begingroup
\begin{multline*}
\Phi (t_{i+1}, t_{i}) 
    = \\ 
      \setlength\arraycolsep{0.5pt}
      \begin{bmatrix}
        \hat{\textbf{R}}^{I_{i+1}} _{I_i} & 0_{3} & 0_{3} &  -\hat{\textbf{R}}^{I_{i+1}} _{I_i} \textbf{J}_r(^{I_{i+1}} _{I_i} \hat{\bm{\theta}}) \Delta t & 0_{3}\\
        -\frac{1}{2} \hat{\textbf{R}}^{G}_{I_i}[\hat{\textbf{a}}_i \Delta t ^2]_{\times} & I_3 & I_3 \Delta t & 0_3 & -\frac{1}{2} \hat{\textbf{R}}^{G}_{I_i} \Delta t ^{2} \\
        -\hat{\textbf{R}}^{G}_{I_i} [\hat{\textbf{a}}_{i} \Delta t]_{\times} & 0_{3} & I_{3} & 0_{3} & -\hat{\textbf{R}}^{G}_{I_i} \Delta t \\
        0_{3} & 0_{3} & 0_{3} & I_{3} & 0_{3} \\
        0_{3} & 0_{3} & 0_{3} & 0_{3} & I_{3}
      \end{bmatrix}
\end{multline*}
\endgroup
\begin{equation*}
    \textbf{G}_{i} 
    = \begin{bmatrix}
    - \hat{\textbf{R}}^{I_{i+1}} _{I_{i}} \textbf{J}_{r} (^{I_{i+1}} _{I_{i}} \bm{\theta}) \Delta t & 0_{3} & 0_{3} & 0_{3} \\
     0_{3} & -\frac{1}{2}\hat{\textbf{R}}^{G}_{I_i} \Delta t^{2} & 0_{3} & 0_{3} \\
     0_{3} & -\hat{\textbf{R}}^{G}_{I_i} \Delta t & 0_{3} & 0_3 \\
     0_{3} & 0_{3} & I_{3} & 0_{3} \\
     0_{3} & 0_{3} & 0_{3} & I_{3}
      \end{bmatrix}
\end{equation*}

Where, $\hat{\textbf{R}}^{I_{i+1}} _{I_i} = \exp(-\hat{\bm{\omega}}_{i} \Delta t)$, $^{I_{i+1}} _{I_i} \hat{\bm{\theta}} = -\hat{\bm{\omega}}_{i} \Delta t$ and $\textbf{J}_r(\bm{\theta})$ is the right Jacobian of $SO(3)$ that maps the variation of rotation angle in the parameter vector space into variation in the tangent vector space to the manifold \cite{barfoot_2017}.
$\textbf{Q}_{d}$ is the IMU noise covariance matrix which can be computed as done in \cite{Trawny2005IndirectKF} (Equations 129-130 $\&$ Equations 187-192). Computation of $\textbf{Q}_d$ requires the knowledge of IMU intrinsic calibration parameters, \emph{viz.} gyroscope/accelerometer noise densities and random walk (in-run biases), which can be looked up in the IMU data-sheet or determined using tools available online\footnote{\url{https://github.com/rpng/kalibr_allan}}. 

\subsubsection{Deskewing Scan}
\label{sec: deskewing}  
The 3D Lidar sequentially produces point measurements using a rotating mechanism. When the Lidar moves, the raw scan produced by it suffers from motion distortion. The calibration data collection process (Section \ref{sec: datacollection}) requires the sensor suite to exhibit motion excitation, which moves the points in a raw scan away from their true positions. In a Lidar scan, each 3D point is measured from a temporally unique frame and comes with a timestamp (which is somewhere between two adjacent scan timestamps). In order to address the problem of motion distortion, we need to predict the IMU pose at point timestamp. The IMU propagation model (Equation \ref{eqn: propeqn1} - \ref{eqn: propeqn7}) is used for IMU pose prediction at point timestamp. Once we have an estimate of the IMU pose at point timestamp, we use the best known estimate of the extrinsic calibration parameter $T^{I}_{L}$ to infer the corresponding Lidar pose, which can be done by exploiting the motion constraint given in Equation \ref{eqn: fullmotionconstraint}. For example, consider a point $x^{L}_{k_i}$ in the $k^{th}$ scan bearing timestamp $k_i$. In order to deskew this point, we manipulate Equation \ref{eqn: fullmotionconstraint} and use it to estimate Lidar motion $T^{L_k}_{L_{k_i}}$ between the scan timestamp $k$ and the point timestamp $k_{i}$ (Equation \ref{eqn: deskewing})
\begin{align}
    \label{eqn: deskewing}
    T^{L_k}_{L_{k_i}} = (\hat{T}^{I}_{L})^{-1}(\hat{T}^{G}_{I_k})^{-1} \hat{T}^{G}_{I_{k_i}}\hat{T}^{I}_{L}
\end{align}
$T^{L_k}_{L_{k_i}}$ calculated in Equation \ref{eqn: deskewing} is used to transform the point $x^{L}_{k_i}$ for obtaining a deskewed Lidar scan. Here, $\hat{T}^{I}_{L}$ is the best known estimate of extrinsic calibration $T^{I}_{L}$ at that instant, $\hat{T}^{G}_{I_k}$ is an estimate of IMU pose at scan timestamp $k$ and finally $\hat{T}^{G}_{I_{k_i}}$ is the pose of the IMU at point timestamp $k_i$, obtained using the IMU state propagation model (Equation \ref{eqn: propeqn1} - \ref{eqn: propeqn7})\footnote{Check the effect of deskewing in the video we have posted here \url{https://youtu.be/YmTyoQA4NaY}}.
\subsubsection{NDT Scan Matching}
\label{sec: ndtscanmatching}
After we deskew the scan we use NDT scan matching \cite{NDT} to generate Lidar motion estimates $T^{L_{k-1}}_{L_{k}}$ between consecutive deskewed Lidar scans $k-1$ and $k$. We use these Lidar motion estimates as measurement for the EKF state update.

\subsubsection{State Update}
\label{sec: stateupdate}
The State Update module requires the knowledge of a measurement model, measurement residual and the measurement Jacobians with respect to the state variables. In this section we will present the measurement model and the measurement residual, but we will omit the derivation of measurement Jacobians in the interest of space. As described in the previous section, we use the result of NDT scan matching as measurement which is parameterized as $T^{L_{k-1}}_{L_{k}}$. We will use the motion constraint in Equation \ref{eqn: fullmotionconstraint} to derive our measurement model. Manipulating Equation \ref{eqn: fullmotionconstraint} gives us the measurement model (Equation \ref{eqn: measurementmodel}):

\begin{align}
    T^{L_{k-1}}_{L_{k}} &= (T^{I}_{L})^{-1} (T^{G}_{I_{k-1}})^{-1} T^{G}_{I_{k}}T^{I}_{L}\label{eqn: measurementmodel} 
\end{align}
The LHS of Equation \ref{eqn: measurementmodel} is the measurement and the RHS is a function of state variables $T^{I}_{L}$, $T^{G}_{I_{k-1}}$, $T^{G}_{I_{k}}$. So, the measurement model (Equation \ref{eqn: measurementmodel}) is in agreement with the standard form $z = h(x)$ used in EKF, where $z$ is the measurement and $h()$ is the measurement model which is a function of the state $x$. In our case, measurement $z=T^{L_{k-1}}_{L_{k}}$ $\&$ measurement model $h(x)=(T^{I}_{L})^{-1} (T^{G}_{I_{k-1}})^{-1} T^{G}_{I_{k}}T^{I}_{L}$ and state $x = \{ T^{G}_{I_{k-1}}, T^{G}_{I_{k}}, T^{I}_{L}\}$.
Here, 

\begin{align*}
    T^{I}_{L} = 
    \begin{bmatrix}
    \textbf{R}(^{I}_{L}\Bar{q}) &  ^{I}\textbf{p}_{L}\\
    0 & 1
    \end{bmatrix}, 
    T^{G}_{I_{k-1}} = 
    \begin{bmatrix}
    \textbf{R}^{T}(^{I_{k-1}} _{G}\Bar{q}) &  ^{G}\textbf{p}_{I_{k-1}}\\
    0 & 1
    \end{bmatrix}
\end{align*} 
\begin{align*}
    T^{G}_{I_{k}} = 
    \begin{bmatrix}
    \textbf{R}^{T}(^{I_{k}} _{G}\Bar{q}) &  ^{G}\textbf{p}_{I_{k}}\\
    0 & 1
    \end{bmatrix}
\end{align*}
Clearly $T^{I}_{L}$, $T^{G}_{I_{k-1}}$, $T^{G}_{I_{k}}$ depend on state variables. 

Separating the rotation and translation components of Equation \ref{eqn: measurementmodel}, we obtain.

\begin{align}
    R^{L_{k-1}}_{L_{k}} &= \textbf{R}^{\top}(^{I}_{L}\Bar{q})\textbf{R}(^{I_{k-1}} _{G}\Bar{q})\textbf{R}^{T}(^{I_{k}} _{G}\Bar{q})\textbf{R}(^{I}_{L}\Bar{q}) \nonumber \\
    {}^{L_{k-1}}p_{L_{k}} &=\textbf{R}^{\top}(^{I}_{L}\Bar{q}) \bigg[\bigg( \textbf{R}(^{I_{k-1}} _{G}\Bar{q})\textbf{R}^{T}(^{I_{k}} _{G}\Bar{q})- \mathbf{I}\bigg) {}^{I}\textbf{p}_{L} \nonumber \\&+ \textbf{R}(^{I_{k-1}} _{G}\Bar{q}) \bigg( {}^{G}\textbf{p}_{I_{k}} - ^{G}\textbf{p}_{I_{k-1}}\bigg) \bigg]
    \label{eqn: measurementmodelseparated}
\end{align}

The measurement model calculated at state estimates gives us the predicted rotation and translation measurement, \emph{viz.} $\hat{R}^{L_{k-1}}_{L_{k}}$ and ${}^{L_{k-1}}\hat{p}_{L_{k}}$ respectively. The difference between the true measurements and predicted measurements gives us the measurement residual $\textbf{r}_k$ required for state update (Equation \ref{eqn: measurementresidual}).

\begin{align}
    \textbf{r}_k = 
    \begin{bmatrix}
    \texttt{Log}(R^{L_{k-1}}_{L_{k}} (\hat{R}^{L_{k-1}}_{L_{k}})^{\top}) \\
    {}^{L_{k-1}}p_{L_{k}} - {}^{L_{k-1}}\hat{p}_{L_{k}}
    \end{bmatrix}
    \label{eqn: measurementresidual}
\end{align}
Here, $\texttt{Log()}$ associates a matrix $R \in SO(3)$ to a vector $\in R^{3\times1}$ (via a skew symmetric matrix). In addition to the measurement residual $\textbf{r}_{k}$, we also require the Jacobians of the measurement model with respect to the state variables in order to perform state and covariance update. The Jacobians (Equation \ref{eqn: Jacobian}) are evaluated at the best available estimate of the state variables $x = \{ T^{G}_{I_{k-1}}, T^{G}_{I_{k}}, T^{I}_{L}\}$. 
\begin{align}
    \textbf{H}^{T^{L_{k-1}}_{L_{k}}}_{T^{I}_{L}} &= \frac{\partial T^{L_{k-1}}_{L_{k}}}{\partial T^{I}_{L}}\bigg|_{\hat{x} = \{ \hat{T}^{G}_{I_{k-1}}, \hat{T}^{G}_{I_{k}}, \hat{T}^{I}_{L}\}} \nonumber\\
    \textbf{H}^{T^{L_{k-1}}_{L_{k}}}_{T^{G}_{I_{k-1}}} &= \frac{\partial T^{L_{k-1}}_{L_{k}}}{\partial T^{G}_{I_{k-1}}}\bigg|_{\hat{x} = \{ \hat{T}^{G}_{I_{k-1}}, \hat{T}^{G}_{I_{k}}, \hat{T}^{I}_{L}\}} \nonumber\\
    \textbf{H}^{T^{L_{k-1}}_{L_{k}}}_{T^{G}_{I_{k}}} &= \frac{\partial T^{L_{k-1}}_{L_{k}}}{\partial T^{G}_{I_{k}}}\bigg|_{\hat{x} = \{ \hat{T}^{G}_{I_{k-1}}, \hat{T}^{G}_{I_{k}}, \hat{T}^{I}_{L}\}} 
    \label{eqn: Jacobian}
\end{align}
These individual Jacobians are stacked together to form a consolidated Jacobian $\textbf{H}_{k}$ and used for state update when a measurement update is available. The state update equations are presented in Equations \ref{eqn: updateeqn1}-\ref{eqn: updateeqn3}.
\begin{align}
    \textbf{K}_{k} &= \textbf{P}_{k_{-}}\textbf{H}^{\top}_{k}(\textbf{H}_{k}\textbf{P}_{k_{-}}\textbf{H}^{\top}_{k} + \textbf{R})^{-1} \label{eqn: updateeqn1}\\
    \begin{bmatrix}
    \hat{X}^{G}_{I_{k_{+}}} \\
    \hat{T}^{I}_{L+}
    \end{bmatrix} &=
    \begin{bmatrix}
    \hat{X}^{G}_{I_{k_{-}}} \\
    \hat{T}^{I}_{L-}
    \end{bmatrix} \oplus \textbf{K}_{k}\textbf{r}_{k} \label{eqn: updateeqn2} \\
    \textbf{P}_{k_{+}} &= \textbf{P}_{k_{-}} - \textbf{K}_{k}\textbf{H}_{k} \textbf{P}_{k_{-}} \label{eqn: updateeqn3}
\end{align}
Here $\textbf{K}_{k}$ is the Kalman gain which is used in Equations \ref{eqn: updateeqn2} and \ref{eqn: updateeqn3} for state and state covariance update respectively. \textbf{R} is the tunable measurement covariance matrix. `-' denotes the estimate before update while `+' denotes the estimate after update. $\oplus$ in Equation \ref{eqn: updateeqn2} refers to generic composition which can be algebraic addition for variables on vector space or rotation composition for variables on $SO(3)$.
\section{System Description}
\label{sec:sysdesc} 
Our system (Figure \ref{fig:LidarIMUSystem}) consists of an Ouster 128 Channel Lidar and a Vectornav VN-300 IMU. The Lidar outputs scans at 10 Hz and IMU outputs gyroscope and accelerometer measurements at 400 Hz.

\section{Experiments and Results}
\label{sec: ExperimentsAndResults}
We use our algorithm to calibrate the sensor suite shown in Figure \ref{fig:LidarIMUSystem}. Since, we use an Ouster 128 channel Lidar, in addition to using it in 128 channel mode, we can reconfigure it to 16, 32, 64 channel modes as well. Therefore, we also present the performance of our algorithm when the Lidar is used with lower number of channels/rings, with an aim to establish that our algorithm will give comparative performance irrespective of the number of channels in the Lidar. We collect two different datasets 1 $\&$ 2 by placing the IMU at two different locations \emph{viz.} locations 1 $\&$ 2 (see Figure \ref{fig:LidarIMUSystem}) respectively. In the following sections, all the Figures have been generated using dataset 2 with the 3D-Lidar operating in 128 channel mode. We obtained similar results with all the channel modes, and for both the datasets, but we omit those here in the interest of space. 
\subsection{Convergence}
\begin{figure}[ht!]
\centering
\includegraphics[width=0.5\textwidth]{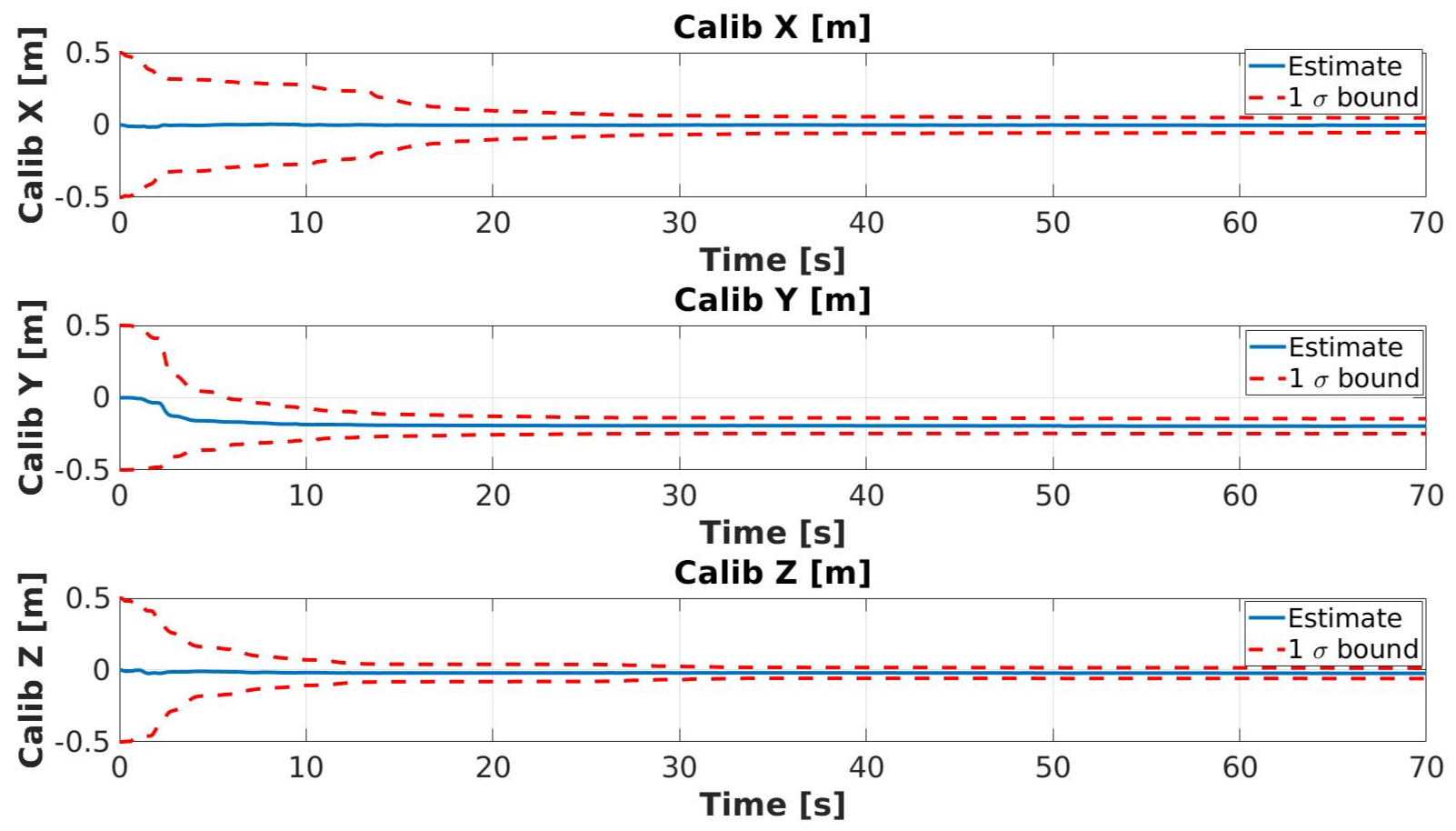}
\caption{Calibration Results for translation component $^{I}p_{L}$ and 1 $\sigma$ bounds. The filter starts from $^{I}\hat{p}_{L} = [0, 0, 0]$ and converges to a fixed value.}
\label{fig: calibXYZfar1}
\vspace{-15pt}
\end{figure}

\begin{figure}[ht!]
\includegraphics[width=0.5\textwidth]{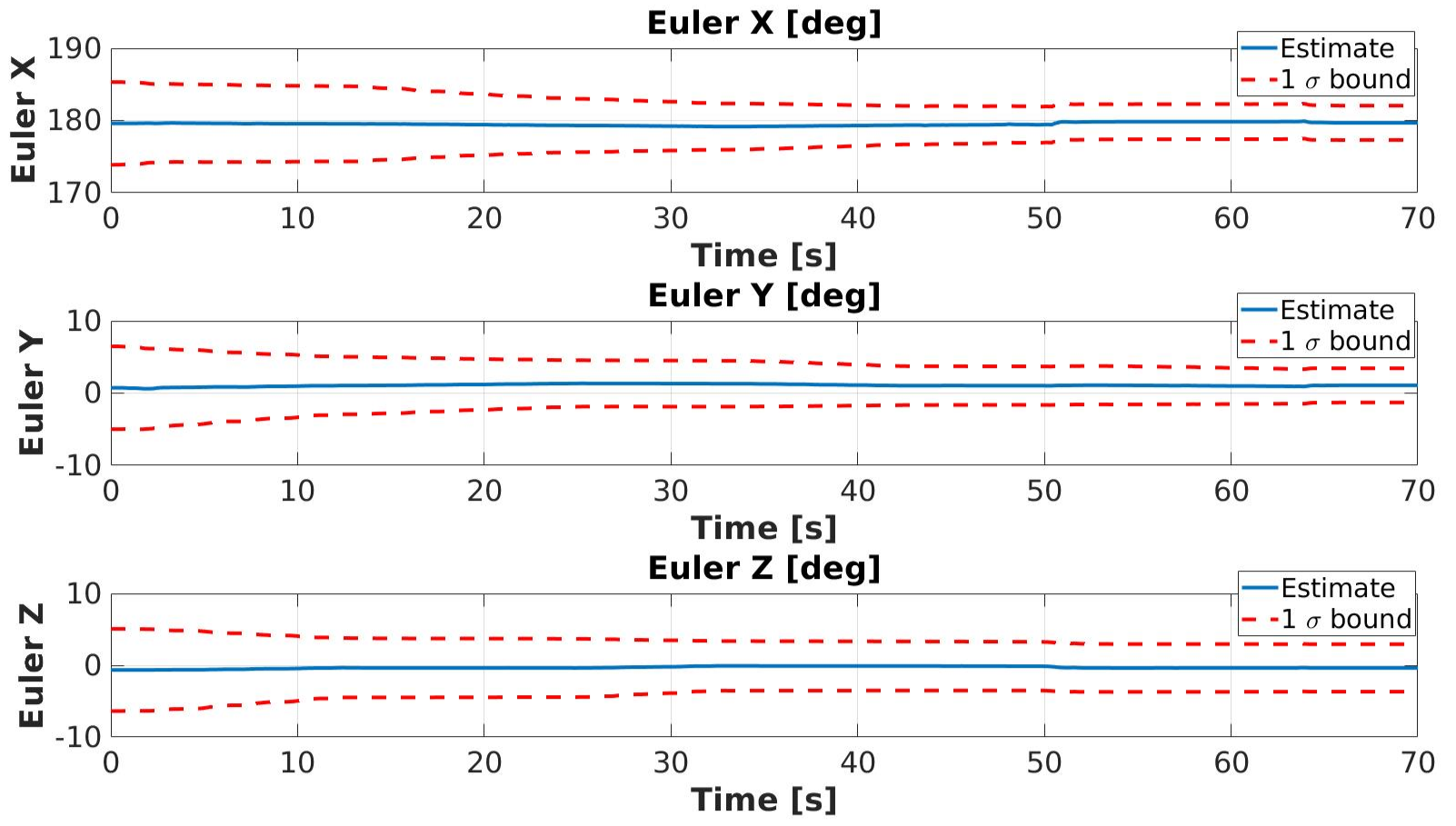}
\caption{Calibration Results for rotation component and 1 $\sigma$ bounds. The filter starts from initialization values that initial rotation component (Section \ref{sec: rotationestimation}) estimation reports.}
\label{fig: calibRotfar1}
\vspace{-20pt}
\end{figure}

Figures \ref{fig: calibXYZfar1} and \ref{fig: calibRotfar1} show the convergence of the estimated extrinsic calibration. As far as the translation variables are considered, we initialize the filter with $^{I}\hat{p}_{L} = [0, 0, 0]$ and the filter converges to fixed values with a tight $\pm1 \sigma$ bound. As far as rotation variables are considered we initialize the filter with estimates obtained from the rotation initialization technique from Section \ref{sec: rotationestimation}. The convergence of the rotation variables with the EKF is shown in Figure \ref{fig: calibRotfar1}. %Although it is difficult to check the accuracy of the converged result without ground truth data, the results obtained is close to the ruler measured results. 

\subsection{Importance of deskewing Lidar scans}
\begin{figure}[ht!]
\centering
\includegraphics[width=0.5\textwidth]{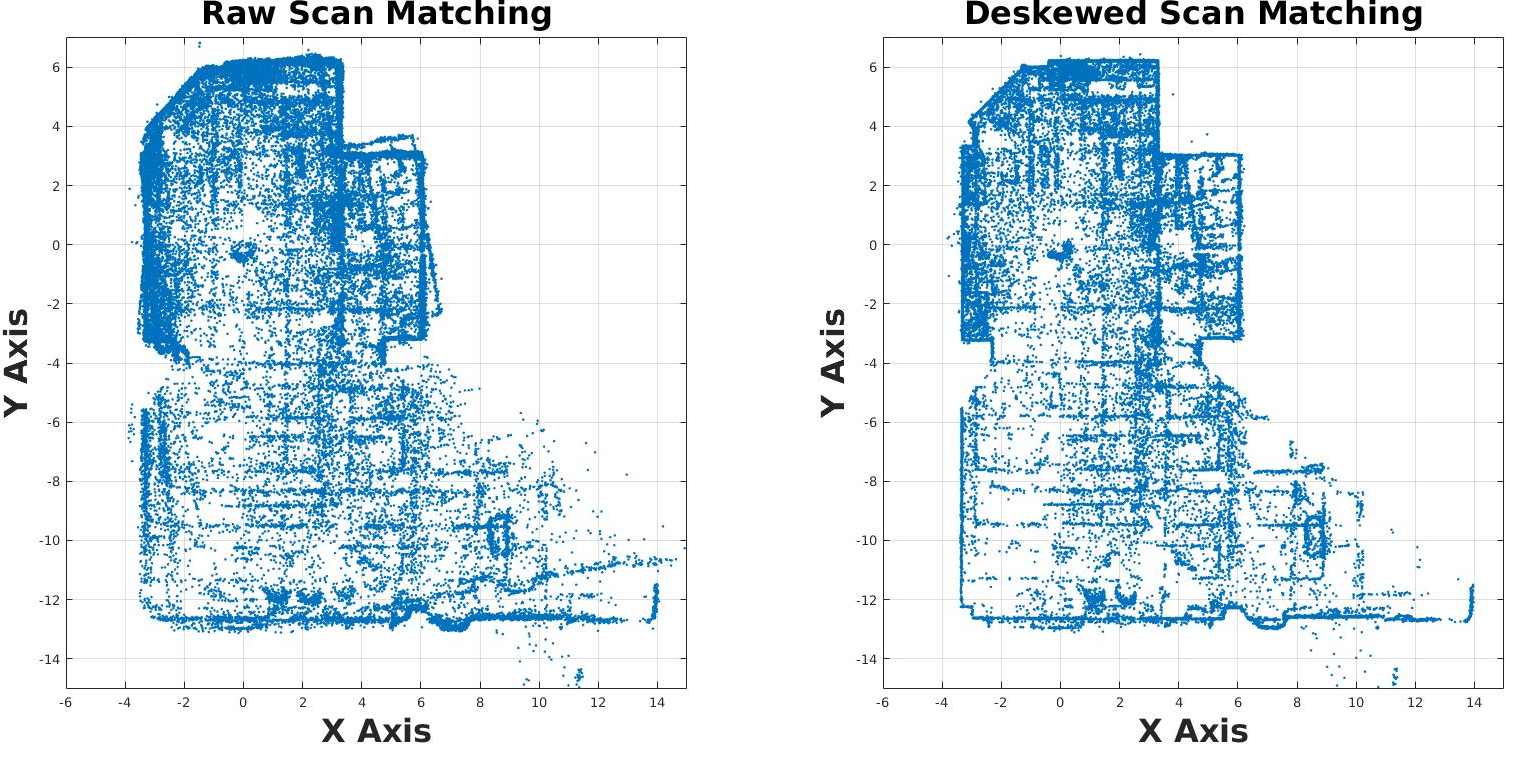}
\caption{Scan Matching with raw and deskewed scans during the calibration process. The scans have been downsampled to improve visibility.}
\label{fig: mappingdeskewvsskew}
\end{figure}

\begin{figure}[ht!]
\centering
\includegraphics[width=0.5\textwidth]{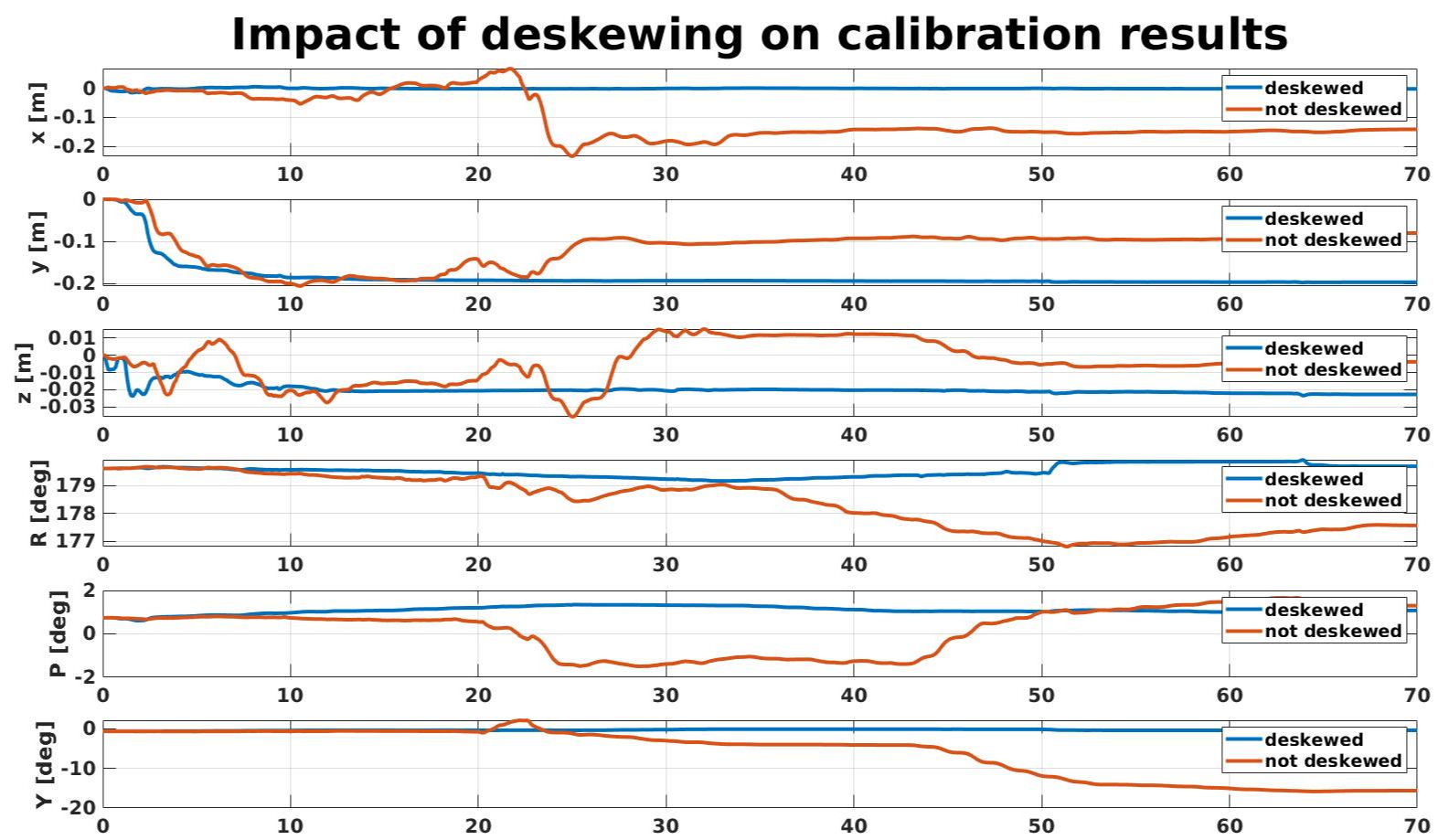}
\caption{The calibration results do not converge to fixed values when the Lidar scan is not deskewed.}
\label{fig: calibdeskewvsskew}
\end{figure}

In this section we present the importance of deskewing the Lidar scans (Section \ref{sec: deskewing}) during the EKF based calibration process (Section \ref{sec: fullstateestimationusingEKF}). In Figure \ref{fig: mappingdeskewvsskew} we present the result of scan matching obtained during the calibration process without and with deskewing of Lidar scans. The left image in Figure \ref{fig: mappingdeskewvsskew} shows blurred and misaligned edges/corners. The edges/corners are blurred because the scan is not deskewed and they are misaligned because the Lidar Odometry from raw scan matching is not correct. Motion distorted Lidar scans lead to inferior results of Lidar Odometry which in turn results in overall deterioration of the calibration estimates (see Figure \ref{fig: calibdeskewvsskew}). Whereas, in the right image of Figure \ref{fig: mappingdeskewvsskew} we see that the edges/corners are sharp and not blurred, thanks to the deskewing of Lidar scans, done using the best available estimate of the extrinsic calibration $\hat{T}^{I}_{L}$ during the EKF based calibration process.  The evolution of calibration results is presented in Figure \ref{fig: calibdeskewvsskew}. The results without deskewing do not converge to fixed values and vary with time when compared against the results obtained when the Lidar scans are deskewed. 
\subsection{Changing Lidar Density}
Tables \ref{table: loc1varryingChannelDensity} and \ref{table: loc2varryingChannelDensity} present the final calibration estimates obtained at the end of the Kalman Filtering process, for different ring numbers (i.e. different channel modes), with datasets 1 $\&$ 2 collected at locations 1 $\&$ 2 (see Figure \ref{fig:LidarIMUSystem}) respectively. In both Tables \ref{table: loc1varryingChannelDensity} $\&$ \ref{table: loc2varryingChannelDensity} we find that our calibration algorithm converges to comparable estimates of calibration parameters irrespective of the number of rings/channels, thus proving its usability across Lidars with varying channel numbers.
\begin{table}[ht!]
\centering
\resizebox{\columnwidth}{!}{
\begin{tabular}{||c || c | c | c | c | c | c||} 
 \hline
 \textbf{No. rings} & \textbf{R}${}^\circ$ & \textbf{P}${}^\circ$ & \textbf{Y}${}^\circ$ & \textbf{x} [m] & \textbf{y} [m] & \textbf{z} [m]\\ 
 \hline\hline
 \textbf{128} & -180.0355 & -0.0956 & -0.7801 & 0.0044 & -0.1455 & -0.0246 \\ \hline
 \textbf{64} & -179.9378 & -0.1337 & -0.8293 & 0.0028 & -0.1458 & -0.0234\\ \hline
 \textbf{32} & -179.9213 & -0.0311 & -1.0018 & 0.001 & -0.1448 & -0.0210 \\ \hline
 \textbf{16} & -179.9129 & 0.1576 & -1.0825 & -0.0035 & -0.1458 & -0.0235 \\\hline
%  $\mu$ & -179.9519 & -0.0257 & -0.9234 & 0.0012 & -0.1455 & -0.0231 \\\hline
%  $\sigma$ & 0.0567 & 0.1293 & 0.1424 & 0.0034 & 0.0004717 & 0.0015 \\[1ex] 
\end{tabular}}
\caption{Results with IMU at Location 1 with the Lidar used with different channel modes}
\label{table: loc1varryingChannelDensity}
\vspace{-20pt}
\end{table}

\begin{table}[ht!]
\centering
\resizebox{\columnwidth}{!}{
\begin{tabular}{||c || c | c | c | c | c | c||} 
 \hline
 \textbf{No. rings} & \textbf{R}${}^\circ$ & \textbf{P}${}^\circ$ & \textbf{Y}${}^\circ$ & \textbf{x} [m] & \textbf{y} [m] & \textbf{z} [m]\\ 
 \hline\hline
 \textbf{128} & -179.6770 & 1.0774 & -0.3735 & -0.0024 & -0.1971 & -0.0227 \\ \hline
 \textbf{64} & -179.1621 & 0.8255 & -0.0392 & -0.0020 & -0.1956 & -0.0248\\ \hline
 \textbf{32} & -179.2619 & 0.8368 & -0.2595 & -0.0031 & -0.1976 & -0.0231 \\ \hline
 \textbf{16} & -179.1540 & 1.1280 & -0.3603 & -0.0041 & -0.2005 & -0.0241 \\\hline
%  $\mu$ & -179.3137 & 0.9669 & -0.2581 & -0.0029 & -0.1977 & -0.0237 \\\hline
%  $\sigma$ & 0.2471 & 0.1582 & 0.1546 & 0.0009201 & 0.0021 & 0.0009535 \\[1ex] 
\end{tabular}}
\caption{Results with IMU at Location 2 with the Lidar used with different channel modes}
\label{table: loc2varryingChannelDensity}
\vspace{-20pt}
\end{table}
\subsection{Relative Verification}
The usual ways to validate calibration algorithms is to use simulators, auxiliary sensors like cameras or GPS, or, in the case of OEMs, CAD diagrams as ground truth. As we assemble our sensor suite ourselves by procuring sensors from different vendors, we do not have access to ground truth. 
% Moreover, using auxiliary sensors like cameras may not be the best way to validate calibration, because cameras are projective sensors, while calibration is a metric problem. Projection of 3D Lidar points on a 2D image plane for verification will not guarantee metric correctness. 
Furthermore, using any auxiliary sensor will also involve an additional overhead of calibrating that sensor w.r.t the IMU. In this work we use a relative verification technique which can validate our calibration results against a reference obtained by ruler measurement. As described earlier, we collect two datasets 1 $\&$ 2 with the IMU kept at two different locations 1 $\&$ 2 respectively (see Figure \ref{fig:LidarIMUSystem}). The relative offset between the two IMU locations is 5 cm along y-axis according to our ruler measurement and we use this as a reference to compare against our calibration results. In order to validate our calibration algorithm, we run it for datasets collected at both of these locations and for all channel modes, and present the difference in the calibration for both the locations in Table \ref{table: far1Andnear1DiffvarryingChannelDensity} (which is the difference between the results presented in Tables \ref{table: loc1varryingChannelDensity} and \ref{table: loc2varryingChannelDensity}).
\begin{table}[ht!]
\centering
\resizebox{\columnwidth}{!}{
\begin{tabular}{||c || c | c | c | c | c | c||} 
 \hline
 \textbf{No. rings} & \textbf{R}${}^\circ$ & \textbf{P}${}^\circ$ & \textbf{Y}${}^\circ$ & \textbf{x} [m] & \textbf{y} [m] & \textbf{z} [m]\\ 
 \hline\hline
 \textbf{128} & 0.3424 & 1.1778 & 0.4061 & -0.0076 & \textcolor{blue}{0.0516} & 0.0019 \\ \hline
 \textbf{64} & 0.5886 & 0.9678 & 0.7887 & -0.0056 & \textcolor{blue}{0.0497} & 0.0014\\ \hline
 \textbf{32} & 0.6442 & 0.8793 & 0.7420 & -0.0050 & \textcolor{blue}{0.0527} & 0.0021 \\ \hline
 \textbf{16} & 0.7404 & 0.9846 & 0.7244 & -0.0016 & \textcolor{blue}{0.0547} & 0.0005 \\ \hline
%  $\mu$ & 0.5789 & 1.0024 & 0.6653 & -0.005 & 0.0522 & 0.0015 \\\hline
%  $\sigma$ & 0.1697 & 0.1257 & 0.1749 & 0.0025 & 0.0021 & 0.0007135 \\[1ex] 
\end{tabular}}
\caption{Difference between calibration results for datasets collected at Location 1 and Location 2 with varying ring density. The difference along y-axis has been highlighted in blue.}
\label{table: far1Andnear1DiffvarryingChannelDensity}
\vspace{-15pt}
\end{table}
Using the aforementioned ruler measurement as reference, we achieve percentage errors of 3.2 $\%$, 0.6 $\%$, 5.4 $\%$ $\&$ 9.4$\%$ along the y-axis for 128, 64, 32 $\&$ 16 channel modes respectively.

\subsection{Motion Required for Observability}
\label{sec: motionandconvergence}
\begin{figure}[ht!]
\includegraphics[width=0.5\textwidth]{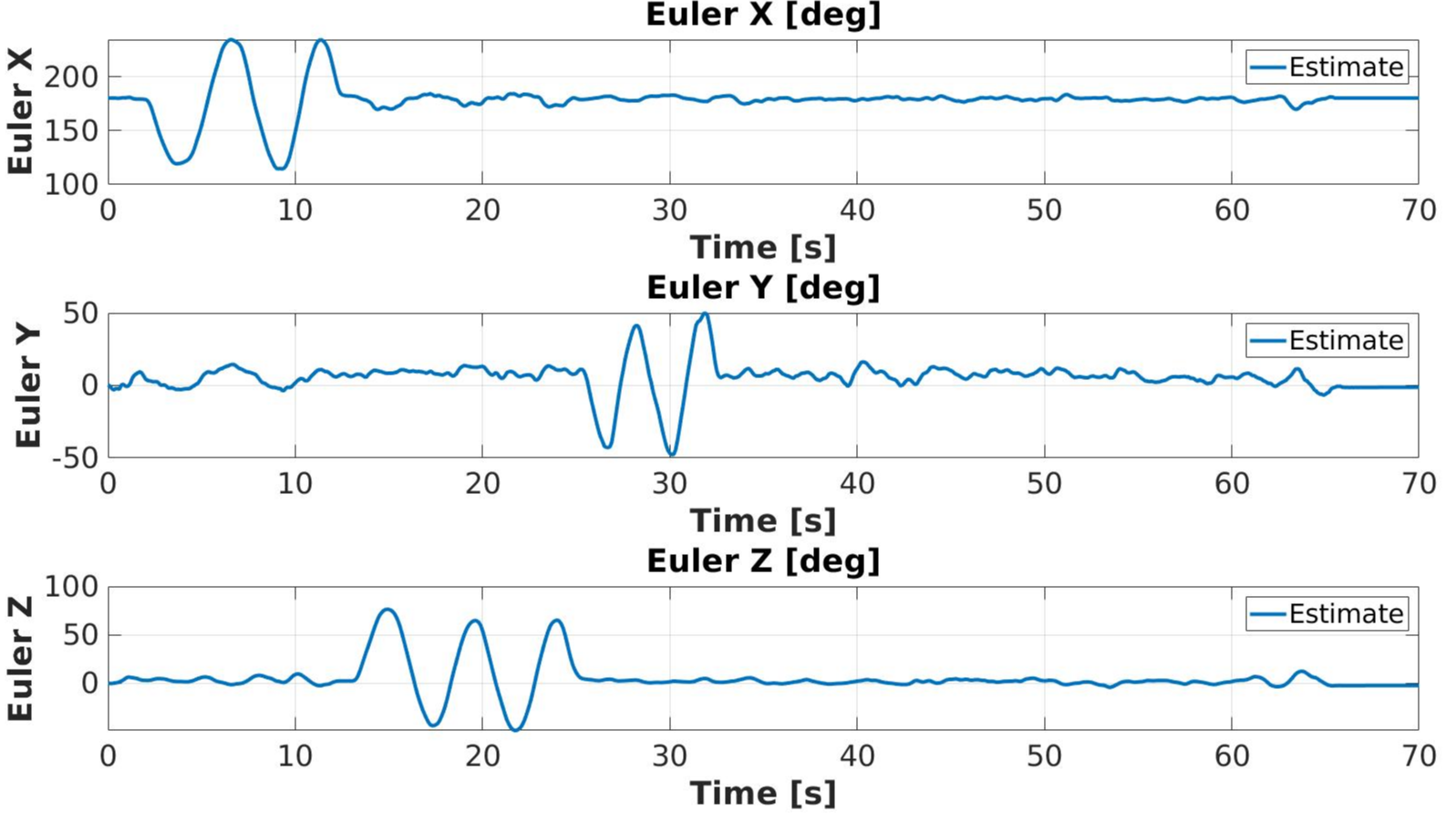}
\caption{Rotation trajectory during data collection}
\label{fig: far1RotTrajectory}
\vspace{-10pt}
\end{figure}

\begin{figure}[ht!]
\centering
\includegraphics[width=0.5\textwidth]{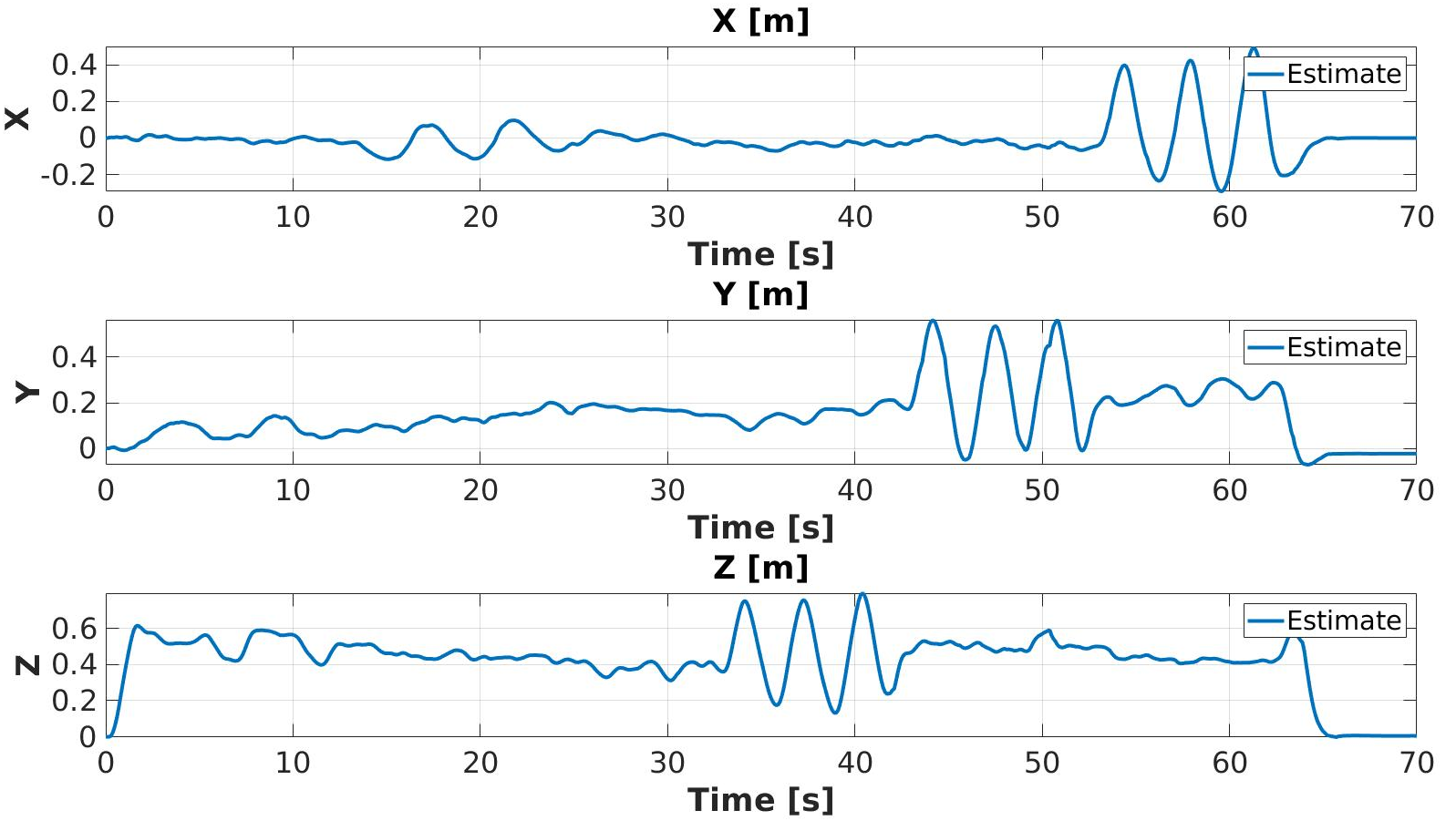}
\caption{XYZ trajectory during data collection}
\label{fig: far1TranslationTrajectory}
\vspace{-20pt}
\end{figure}
As mentioned in \cite{KalmanFilterBasedCamCalib}, it is advisable to excite at least two degrees of rotational freedom for sufficiently long time at the beginning of the calibration process in order to ensure convergence to reliable calibration values. This is demonstrated on comparing Figures \ref{fig: calibXYZfar1} $\&$ \ref{fig: calibRotfar1} which show the convergence of the calibration parameters, against Figures \ref{fig: far1RotTrajectory} $\&$ \ref{fig: far1TranslationTrajectory} which present the rotation and translation trajectory undertaken by the sensor suite during the data collection procedure (Section \ref{sec: datacollection}). In Figure \ref{fig: calibXYZfar1}, the translation calibration parameters converge in 25-30 seconds which correspond to the time when the rotation motions (see Figure \ref{fig: far1RotTrajectory}) about two axes (\emph{viz.} X $\&$ Z) cease to be performed. As far as the rotation calibration parameters are considered (Figure \ref{fig: calibRotfar1}) they vary marginally during the EKF based calibration process (Section \ref{sec: fullstateestimationusingEKF}) as they are already initialized close to final estimates during the initialization procedure (Section \ref{sec: rotationestimation}).
\section{Conclusion}
\label{sec: conclusionAndFutureWork}
In this work, we presented an EKF based 3D-Lidar IMU calibration algorithm which does not depend on any calibration target or auxiliary sensors for estimating the extrinsic calibration. We presented the usability of our algorithm for different Lidar channel modes, and concluded that it gives comparable results for all the channel modes for both the datasets we had collected from two different IMU locations. We also presented a relative way of validating calibration algorithms when we do not have access to ground truth or we do not have an auxiliary sensor at our disposal.
% \section{Acknowledgement}
% The authors would like to thank George Chustz for assembling the sensor suite.
\bibliographystyle{IEEEtran}
\bibliography{bibexpendable}

% Generated by IEEEtran.bst, version: 1.14 (2015/08/26)
\begin{thebibliography}{10}
\providecommand{\url}[1]{#1}
\csname url@samestyle\endcsname
\providecommand{\newblock}{\relax}
\providecommand{\bibinfo}[2]{#2}
\providecommand{\BIBentrySTDinterwordspacing}{\spaceskip=0pt\relax}
\providecommand{\BIBentryALTinterwordstretchfactor}{4}
\providecommand{\BIBentryALTinterwordspacing}{\spaceskip=\fontdimen2\font plus
\BIBentryALTinterwordstretchfactor\fontdimen3\font minus
  \fontdimen4\font\relax}
\providecommand{\BIBforeignlanguage}[2]{{%
\expandafter\ifx\csname l@#1\endcsname\relax
\typeout{** WARNING: IEEEtran.bst: No hyphenation pattern has been}%
\typeout{** loaded for the language `#1'. Using the pattern for}%
\typeout{** the default language instead.}%
\else
\language=\csname l@#1\endcsname
\fi
#2}}
\providecommand{\BIBdecl}{\relax}
\BIBdecl

\bibitem{lincalib1}
C.~{Le Gentil}, T.~{Vidal-Calleja}, and S.~{Huang}, ``3d lidar-imu calibration
  based on upsampled preintegrated measurements for motion distortion
  correction,'' in \emph{2018 IEEE International Conference on Robotics and
  Automation (ICRA)}, 2018, pp. 2149--2155.

\bibitem{lincalib2}
J.~{Lv}, J.~{Xu}, K.~{Hu}, Y.~{Liu}, and X.~{Zuo}, ``Targetless calibration of
  lidar-imu system based on continuous-time batch estimation,'' in \emph{2020
  IEEE/RSJ International Conference on Intelligent Robots and Systems (IROS)},
  2020, pp. 9968--9975.

\bibitem{ppccal}
\BIBentryALTinterwordspacing
G.~Pandey, J.~McBride, S.~Savarese, and R.~Eustice, ``Extrinsic calibration of
  a 3d laser scanner and an omnidirectional camera,'' \emph{IFAC Proceedings
  Volumes}, vol.~43, no.~16, pp. 336--341, 2010, 7th IFAC Symposium on
  Intelligent Autonomous Vehicles. [Online]. Available:
  \url{https://www.sciencedirect.com/science/article/pii/S1474667016350790}
\BIBentrySTDinterwordspacing

\bibitem{pbpccal}
S.~Mishra, G.~Pandey, and S.~Saripalli, ``Extrinsic calibration of a 3d-lidar
  and a camera,'' in \emph{2020 IEEE Intelligent Vehicles Symposium (IV)},
  2020, pp. 1765--1770.

\bibitem{msgcal}
J.~L. Owens, P.~R. Osteen, and K.~Daniilidis, ``Msg-cal: Multi-sensor
  graph-based calibration,'' in \emph{2015 IEEE/RSJ International Conference on
  Intelligent Robots and Systems (IROS)}, 2015, pp. 3660--3667.

\bibitem{jiang2021calibrating}
P.~Jiang, P.~Osteen, and S.~Saripalli, ``Calibrating lidar and camera using
  semantic mutual information,'' 2021.

\bibitem{KalmanFilterBasedCamCalib}
F.~M. {Mirzaei} and S.~I. {Roumeliotis}, ``A kalman filter-based algorithm for
  imu-camera calibration: Observability analysis and performance evaluation,''
  \emph{IEEE Transactions on Robotics}, vol.~24, no.~5, pp. 1143--1156, 2008.

\bibitem{kalibr}
P.~Furgale, T.~Barfoot, and G.~Sibley, ``Continuous-time batch estimation using
  temporal basis functions,'' \emph{The International Journal of Robotics
  Research}, vol.~34, pp. 2088--2095, 05 2012.

\bibitem{kalibr2}
J.~Rehder and R.~Siegwart, ``Camera/imu calibration revisited,'' \emph{IEEE
  Sensors Journal}, vol.~17, no.~11, pp. 3257--3268, 2017.

\bibitem{lincalib3}
C.~Glennie, ``Calibration and kinematic analysis of the velodyne hdl-64e s2
  lidar sensor,'' \emph{Photogrammetric Engineering `I\&' Remote Sensing},
  vol.~78, pp. 339--347, 04 2012.

\bibitem{lincalib4}
Z.~Taylor and J.~Nieto, ``Motion-based calibration of multimodal sensor
  arrays,'' vol. 2015, 05 2015.

\bibitem{lincalib5}
\BIBentryALTinterwordspacing
A.~Geiger, P.~Lenz, C.~Stiller, and R.~Urtasun, ``Vision meets robotics: The
  kitti dataset,'' \emph{The International Journal of Robotics Research},
  vol.~32, no.~11, pp. 1231--1237, 2013. [Online]. Available:
  \url{https://doi.org/10.1177/0278364913491297}
\BIBentrySTDinterwordspacing

\bibitem{lincalib6}
J.~Rehder, P.~Beardsley, R.~Siegwart, and P.~Furgale, ``Spatio-temporal laser
  to visual/inertial calibration with applications to hand-held, large scale
  scanning,'' in \emph{2014 IEEE/RSJ International Conference on Intelligent
  Robots and Systems}, 2014, pp. 459--465.

\bibitem{GPR}
C.~Rasmussen and H.~Nickisch, ``Gaussian processes for machine learning (gpml)
  toolbox,'' \emph{Journal of Machine Learning Research, v.11, 3011-3015
  (2010)}, vol.~11, 11 2010.

\bibitem{DBLP:journals/corr/ForsterCDS15}
\BIBentryALTinterwordspacing
C.~Forster, L.~Carlone, F.~Dellaert, and D.~Scaramuzza, ``On-manifold
  preintegration theory for fast and accurate visual-inertial navigation,''
  \emph{CoRR}, vol. abs/1512.02363, 2015. [Online]. Available:
  \url{http://arxiv.org/abs/1512.02363}
\BIBentrySTDinterwordspacing

\bibitem{OpenVins}
P.~{Geneva}, K.~{Eckenhoff}, W.~{Lee}, Y.~{Yang}, and G.~{Huang}, ``Openvins: A
  research platform for visual-inertial estimation,'' in \emph{2020 IEEE
  International Conference on Robotics and Automation (ICRA)}, 2020, pp.
  4666--4672.

\bibitem{mocalZachTaylor}
Z.~Taylor and J.~Nieto, ``Motion-based calibration of multimodal sensor
  extrinsics and timing offset estimation,'' \emph{IEEE Transactions on
  Robotics}, vol.~32, no.~5, pp. 1215--1229, 2016.

\bibitem{hec}
\BIBentryALTinterwordspacing
R.~Horaud and F.~Dornaika, ``Hand-eye calibration,'' \emph{The International
  Journal of Robotics Research}, vol.~14, no.~3, pp. 195--210, 1995. [Online].
  Available: \url{https://doi.org/10.1177/027836499501400301}
\BIBentrySTDinterwordspacing

\bibitem{NDT}
P.~Biber and W.~Straßer, ``The normal distributions transform: A new approach
  to laser scan matching,'' vol.~3, 11 2003, pp. 2743 -- 2748 vol.3.

\bibitem{ceres-solver}
S.~Agarwal, K.~Mierle, and Others, ``Ceres solver,''
  \url{http://ceres-solver.org}.

\bibitem{Trawny2005IndirectKF}
N.~Trawny and S.~Roumeliotis, ``Indirect kalman filter for 3 d attitude
  estimation,'' 2005.

\bibitem{barfoot_2017}
T.~D. Barfoot, \emph{State Estimation for Robotics}.\hskip 1em plus 0.5em minus
  0.4em\relax Cambridge University Press, 2017.

\end{thebibliography}

\end{document}